\documentclass[9pt,a4paper,twocolumn,twoside]{tau-class/tau}
\usepackage[english]{babel}
\usepackage{verbatim}

\journalname{Technical Report}
\title{Leveraging Large Language Models \\for Information Verification -- an Engineering Approach}

\author[a]{Nguyen Nang Hung}
\author[a]{Nguyen Thanh Trong}
\author[a]{Vuong Thanh Toan}
\author[a]{Nguyen An Phuoc}
\author[a]{Dao Minh Tu}
\author[a]{Nguyen Manh Duc Tuan}
\author[a]{Nguyen Dinh Mau}

\affil[a]{\{hungnn49, trongnt11, toanvt7, phuocna3, tudm8, tuannmd, maund\}@fpt.com, FPT Software}

\professor{Corresponding author: Nguyen Manh Duc Tuan (tuannmd@fpt.com)}

\footinfo{}
\leadauthor{Hung Nguyen et al.}
\begin{abstract}    
    For the ACMMM25 challenge, we present a practical engineering approach to multimedia news source verification, utilizing Large Language Models (LLMs) like GPT-4o as the backbone of our pipeline. Our method processes images and videos through a streamlined sequence of steps: First, we generate metadata using general-purpose queries via Google tools, capturing relevant content and links. Multimedia data is then segmented, cleaned, and converted into frames, from which we select the top-K most informative frames. These frames are cross-referenced with metadata to identify consensus or discrepancies. Additionally, audio transcripts are extracted for further verification. Noticeably, the entire pipeline is automated using GPT-4o through prompt engineering, with human intervention limited to final validation. For FAI personals, 
\end{abstract}

\keywords{Technical Report, Source Verification Challenge}

\begin{document}
		
    \maketitle 
    \thispagestyle{firststyle} 
    \tauabstract 
    % \tableofcontents
    % \linenumbers 
    
%----------------------------------------------------------

\section{Introduction}

\taustart{W}ith the rise of technological advancements, information becomes an abundant resource, which has gradually gone uncontrolled. Consequentially, Trojan-bait, agenda-driven propaganda fills up the internet, sowing distrusts and divisions that may result in devastating repercussions \cite{Kuznetsova2021CovidFakeNews}. To combat such harmful information, ACMMM25-Grand Challenge\footnote{https://multimedia-verification.github.io} has provided a platform for news-verification solutions. This report presents our practical approach from an engineering perspective, with the objective of verifying, cross-checking the information, and also minimizing human efforts throughout the process.

\noindent As inspired by existing approaches \cite{Vykopal2024GenerativeLLMs, Zhou2019FakeNewsSurvey}, many of which focus on single modalities or require extensive manual intervention \cite{Yang2023FakeNewsDetectionTech}, we are motivated to design an LLM pipeline as the main part of our solution. Such a pipeline consists of image-processing, audio-processing procedures, and prompt engineering to maximize the coherence of outputs. Our pipeline is powered and automated by GPT-4o API -- which is much more limited compared to its web-service counterpart, hence our challenges:
\begin{enumerate}
    \item {\bf How to effectively process multimedia data?} \\
    It is not trivial to feed a video-content to GPT via its API, thus requiring techniques to transform video data to conventional form (images, texts). For this, we propose a procedure to (i) analyze video data frame-by-frame using the K-mean algorithm \cite{Goyal2014VideoSummarizationKMeans}, then (ii) select representative frames as input for GPT API. As we will show in Section 2, this method is consistent with the workflow of GPT web-service.
    \item {\bf How to verify multimedia data?} \\
    It remains challenging for even human readers to correctly verify the source information since it requires double-check, cross-verify across a wide variety of online resources. This, however, is an advantage of an automated pipeline: we use a Google query API to search for keywords regarding the news. These keywords are extracted in advance from the title or description of the news. As we have acquired an array of internet sources, we summarize their contents, and leverage GPT to cross-verify among internet sources, as well as between internet sources and multimedia contents \cite{Kotonya2024EvaluatingLLMAgents}.
\end{enumerate}

\begin{figure}[t]
    \centering
    \includegraphics[width=\linewidth]{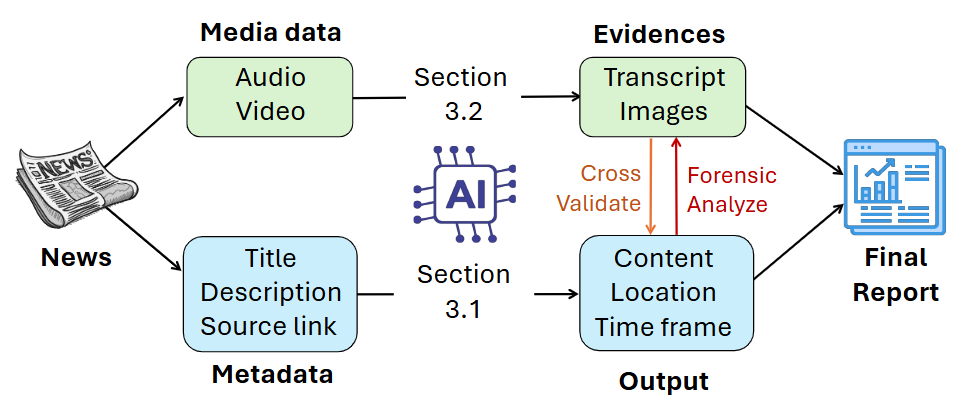}
    \caption{Pipeline overview.}
    \label{fig:overview}
\end{figure}

\noindent The structure of this report is as follows. Section 2 gives a thorough analysis on the data of the Challenge. We also conduct a case study to examine the potential of Large Language Models, particularly GPT-4o, in solving the problem at hands. This case study motivates us to develop companion tools to support GPT-4o as our backbone knowledge base. Section 3 details our pipeline and its main components: multimedia processing and cross verification. Section 4 is dedicated to practical implementations where code blocks and sample outputs are provided as demonstration. We finalize our report with discussion and conclusion in Section 5.

\section{Data Analysis \& Case Study}
\subsection{Input format, output requirements}
As provided by the Challenge, the dataset used for testing contains a total of $50$ samples, each has multimedia data files (videos in mp4 format, or images in jpg format). In which, audio, video data takes up $72.56\%$ of the total data while the other $27.54\%$ being images. It is clear that video-processing is the key for performance in this Challenge.

\noindent As for the output, the three most significant elements are location (geological coordinates), time of event (dates), and evidences. In which, evidences must be cross-checked by both provided multimedia data, and internet sources. It is worth noting that the ultimate goal is not to prove the news" fidelity, but to the largest extend, provide evidences regarding which sections are verifiable by external sources, and which are not.

\subsection{A bird-eye view on task difficulty}
A thorough investigation on the provided output samples shows that $70\%$ of the data is geologically verifiable whereas that for time verifiability is only $30\%$. This stems from the nature of the data itself as its content mostly relates to warfare, conflicts, and violence. Indeed, statistics show over $90\%$ of the data points are related to attacks and explosions. The most prominent locations in data are hospitals and other constructions in Gaza and Ukraine. In general, the data is considered sensitive with high level of violence.

\noindent Due to this level of sensitivity, using policy-regulated LLMs becomes a challenge itself as the generated content may be susceptible to guardrail, e.g., ChatGPT policy. Nonetheless, we find that LLMs, thus far, provide the most flexible tool for reasoning, dialectic, and summarization, which are required for forensic analysis. As for some rare policy-violated cases, such as ones with videos or images containing casualties, we require human interventions for such cases. 

\subsection{A case study}
As a common engineering practice, we first investigate the potential of the most obvious and available tool, i.e., Large Language Models, to examine its capacity for the task. The data point ``ID221"" is chosen to be case-studied. We proceed the test as follows: we upload the entire media data (a minute and a half long video) to GPT-4o web-service, and prompt for summarization of the content. Much to our surprise, GPT"s response has all the required information, as shown below.

\begin{tauenv}[frametitle=GPT-4o Response]
The video is about an Israeli airstrike on a UN-run school in the Nuseirat refugee camp in Gaza.
Here"s a summary of the content:
\begin{itemize}
    \item The school was operated by UNRWA (United Nations Relief and Works Agency) and was sheltering thousands of displaced Palestinians at the time of the strike.
    \item At least 16 people were killed and over 50 injured, according to local health authorities.
    \item The Israeli military (IDF) claimed it targeted Hamas militants using the school as a base for operations and said it took steps to minimize harm to civilians.
    \item Hamas denied the presence of militants at the site and condemned the strike as a "brutal crime" against displaced people.
    \item The UN Secretary-General António Guterres responded with concern, emphasizing that UN premises are inviolable and must be protected.
\end{itemize}
\end{tauenv}

\noindent It is worth noting that without any other external knowledge, GPT-4o web-service is capable to recognize the location, people or grouped involved, and the time frame of the event. Furthermore, a close examination on the media data shows no clear indication in term of text or audio transcript that suggests the location nor involved individuals. 

\noindent To understand its ability to respond such an information-rich output, we proceed to examine how GPT-4o processes media contents. A prompt-engineering approach shows GPT-4o"s procedure:
\begin{enumerate}
    \item Media data is chunked into fixed-sized partitions.
    \item An image is extracted per partition, its content is inferred and put in context with other images from other partitions.
    \item An overall assessment, likely with implemented search tools, concludes that the images from the media match the "airstrike on an UN-affiliated school" event.
    \item A report, probably from multiple internet resources or from its training data, about the event is generated.
\end{enumerate}
\noindent {\bf Limitations.} It is clear that LLMs, such as GPT-4o, are capable of solving the verification task up to the level of inference and reasoning. However, as we aim to develop a fully automated pipeline, only its API is available. This poses a great downside: APT-4o API lacks of tools and reasoning prompts. Furthermore, it is unreasonable to chunk-and-extract images from media data as valuable information may distribute non-uniformly along the length of the media data. Thirdly, GPT-4o dismiss information verification -- which is the main target of the task.

\noindent {\bf Technical Contributions.} First, by recognizing the potential of GPT-4o, we implement a fully automated pipeline with GPT-4o API as the core. Second, since API version lacks prompting and tools compared to its web-service counterpart, we develop the following supplementary supports: (i) a procedure to segment, evaluate, and select information rich frames from the given media data; (ii) a procedure to query online sources, summarize and reason their consensus, as well as their discrepancies; (iii) these information, along with the transcribed audio, are again refined by GPT with task-specific prompting to output the final report. Compared to the web-service version, our report has high coherence, better reasoning, and in-depth cross-validation between information sources.

\section{Solution Specification}
\subsection{Metadata Processing}
\label{subsec:meta}
Metadata includes location, violence level, title, social media link, description, and category. However, most of them are inconsistent, meaning they are not reliable as a consistent element participating in the processing pipeline. As such, we choose three fields to feed the procedure: title, description, and social media link (source link). It is worth noting that source link are mostly either not provided or inaccessible (depending on the region and sensitivity). Nonetheless, since the source link is the first and easiest source of verification, we are reluctant to include this information.

\noindent With the search engine, we select the top-$K$\footnote{Unless otherwise stated, $K$ is default set at $10$.} most viewed results returned by Google. In which, exact matches are of higher priority than the rest. The search keywords are extracted from the description and title of the news. As top-$K$ results are identified (articles, videos, or reports), we crawl text-based content from the internet via the resulted links, and store them locally in buffer. Moreover, event-related information are also extracted from these sources, which include locations, dates, and activities. These information may differ with respect to different sources, thus requiring consensus analysis via cross verification (Section~\ref{subsec:cross-verify}). Hereafter, whenever the term "metadata" is used, it is referred to the processed output from the search engine rather than the original metadata as shown in Figure~\ref{fig:overview}.

\subsection{Video \& Image Processing}
Given that video content constitutes the majority of the challenge dataset (72.56 \% of total data), efficient and intelligent video processing is paramount for our solution. A single news case may contain multiple videos, and each video, especially if it"s a news report, can encompass diverse scenes or segments. Direct input of entire video files into Large Language Models (LLMs) like GPT-4o via API is not feasible due to inherent limitations in their token context window, significant computational overhead, and overall resource constraints. As a standard practice for preparing images for API input, Base64 encoding is commonly utilized. LLM APIs are typically designed to transmit data as text strings, and since images are binary data, Base64 encoding converts this image data into an ASCII string format. This makes it compatible with text-based transmission and allows images to be included as part of an API request, enabling the LLM to effectively receive and process the visual information. However, transmitting every frame as a Base64-encoded image would lead to a enormous large number of tokens, quickly exceeding both practical and economic limits. Therefore, our pipeline employs a multi-step approach to transform raw multimedia into a streamlined format that overcomes these challenges of token limitations, computational overhead, and resource inefficiency. This transformation is crucial for ensuring the data is suitable for LLM analysis while preserving critical information. 

To effectively handle the diverse modalities and inherent complexities of multimedia content, our approach is systematically divided into three main components: Video Frame Extraction and Selection, Audio Transcription and Image Processing.

\subsubsection{Video Frame Extraction and Selection}
For video data, the primary challenge lies in effectively reducing the visual information to a manageable set of key frames without losing critical context. Our approach addresses this by first segmenting the video into meaningful units and then selecting representative frames from these units.

\noindent{\bf Shot Detection and Segmentation.} The video is initially processed using TransNetV2, a pre-trained shot detection model \cite{Soucek2020TransNetV2}. This deep neural network builds upon its predecessor, TransNet, primarily utilizing Dilated Deep Convolutional Neural Network (DCNN) cells as its backbone architecture \cite{Soucek2020TransNetV2}. These cells incorporate 3D convolution operations with various dilation rates, enabling the model to effectively process sequences of frames and capture temporal dependencies for accurate shot transition detection. This model analyzes the video frame by frame to identify significant visual transitions, effectively segmenting the video into distinct "shots." A shot represents a continuous, uninterrupted sequence of frames, often conveying a single event or scene within the larger video. This is crucial because a news report video might feature multiple distinct scenes that need to be individually represented. The model"s output is then used to define the start and end frames for each detected shot.

\noindent{\bf Feature Extraction per Shot.} For each detected shot, frames are extracted. Visual features from these frames are then obtained using a pre-trained Vision Transformer (ViT) model \cite{Dosovitskiy2020ViT}. ViT architectures are particularly effective for this task as their self-attention mechanism allows the model to integrate information globally across the entire image, which is valuable for understanding the overall scene in a news frame \cite{vit_papersummary}. This model generates high-dimensional embeddings that numerically represent the content of each frame within the shot. These embeddings form the basis for subsequent clustering.

\noindent{\bf Representative Key-frames Selection.} To identify the most informative frames within each shot and reduce redundancy across the entire video, we apply the $K$-means clustering algorithm to the extracted frame features for each detected shot \cite{HuynhLam2023TACSUM}. Each cluster centroid within a shot represents a distinct visual moment or event within that shot. From each cluster, the frame closest to its centroid (or a set of frames) is selected as a "key frame." The optimal number of clusters for each shot is determined by evaluating clustering quality metrics, the silhouette score, over a predefined range of possible cluster counts. This method ensures that the selected top-$K$ frames collectively represent the diverse visual content of the entire video by covering all its significant shots. This strategic selection is crucial, as valuable information often distributes non-uniformly along the video's timeline.

\noindent{\bf Cross-Video Key-frames Aggregation.} If a single news case involves multiple videos, the representative frames identified from each individual video"s shots are collected. These aggregated frames are then subjected to a final round of $K$-means clustering to select an overall, globally representative set of frames for the entire case. This ensures that the most unique and informative visual content across all provided videos is prioritized for LLM analysis, further optimizing token usage.

\noindent{\bf Downscaling and Preparation for LLM Input:} The final set of selected key frames are then downscaled to a resolution suitable for API input, balancing visual fidelity with token limits (e.g., to 256x256 pixels). These images are then encoded into Base64 format for efficient submission to the GPT-4o API for visual analysis.

\subsubsection{Audio Transcription}
In parallel with video frame processing, audio tracks embedded within the video content are extracted. To prepare this audio for transcription, the stream is segmented into 30-second chunks. These chunks are then processed using OpenAI's Whisper model to transcribe speech into text \cite{Radford2023Whisper}. The Whisper model is adept at handling various languages, a crucial feature for the challenge's diverse dataset \cite{Radford2023Whisper}. It can be set to either auto-detect the language or be configured for a specific one, such as Arabic. The resulting audio transcripts provide an additional layer of textual information that can be cross-referenced with visual content and external metadata. 

\subsubsection{Image Processing}
For static image inputs, the processing is more straightforward. Single images are directly prepared for LLM input by ensuring they are appropriately encoded. In cases where multiple images are provided for a single news event, a similar feature extraction and $K$-means clustering approach is applied to select the most representative or distinct images, preventing redundant information from being fed to the LLM and optimizing API usage.

\subsection{Information Verification}
\label{subsec:cross-verify}
\noindent{\bf Cross Validation.} It is essential that a news verification system can cross validate its sources. As we have successfully retrieved and process metadata (online articles and other information sources -- Section~\ref{subsec:meta}), we force GPT to compare, contrast, and condense the information. Again, we leverage prompt engineering as seen in Prompt 1. The targets of validation include the location of event, the time frame of the event, and the content of the event. Among these fields, contradictions often occur for the time frame. As such, we classify the following concepts: \textit{Consensus} (if the difference in time frame of the reported event is less than month), \textit{Partial} (if the difference is from one to three months), and \textit{Non-verifiable} (otherwise). We also seek for discrepancies about the content of these sources. We want to highlight that coordinates, although extracted from GPT based on the location name, match that provided by Google Map with little variations. 
\begin{tauenv}[frametitle=Prompt 1: Cross-validate information sources]
\begin{verbatim}
You are given sources of information about an event, 
and you are asked to summarize them to have an 
overview of the event"s location, date, abouts.

<location>: where is/was the event? 
    Allocate the exact location 
    (road/bridge/building/site) and (region/country) 
    and get the exact coordinates. Format:
    {
    "location": ...,
    "coordinates": ...
    }
<date>: when is/was the event? 
    Choose time frame to encompass all information 
    sources. If the time frame is less than 1 month, 
    it is Consensus. If the time frame is from 
    1 month to 3 months, it is Partial.
    Otherwise, it is Non-verifiable. 
    If non-verifiable, perhaps the event reoccurred, 
    what time frame has the most reported sources?
    Format:
    {
    "date":  "dd/mm/yyyy",
    "concensus": "Yes"/"Partial"/"Non-verifiable",
    "notes": first and lastest occurrence,
    }
<abouts>: what is/was the event about? 
    Compare to find consensus and conflicts 
    between sources. Format:
    {
    "consensus": ...,
    "conflicts": ...
    }
<tags>: Assign relevant tags based on platforms, 
    people, brands, or specific topics 
    (e.g., TikTok, Trump, Coca-Cola, 
    Ukraine War, or AI-generated).

### Output format:
{
    "location": <location>,
    "date": <date>,
    "about": <abouts>,
    "tag": <tags>
}
\end{verbatim}
\end{tauenv}
\noindent{\bf Forensic Analysis.} In this work, we implicitly assume that the information from media data, if found unedited, is the undeniable ground-truth to verify other information. With images ready, we can use this information to verify the output of Prompt 1. Prompt 2 details our desired input and output:
\begin{tauenv}[frametitle=Prompt 2: Forensic Analysis Prompt]
\begin{verbatim}
You will receive images from a video, and metadata. 
You are tasked to:
(1). Find evidences to validate information in the
metadata. Format:
{   
    "location": Describe the location
    (road/bridge/building/site) and get its name.
    Is it the same location in metadata? Evidences?,
    "event": Is it the same event? Evidences?,
    "people": Identify key individuals or groups
}
(2). Check if the content is synthetic, modified.
(3). Base on which evidence that you think it is 
     authentic, synthetic, or modified.
(4). Identify AI-generated content 
     (e.g., GANs, Stable Diffusion).
(5). Note any detected anomalies or manipulations. 
### Output Format
{   
    "metadata-validation": (1),
    "authenticity": (2),
    "auth-evidence": (3),
    "synt-type": (4),
    "other": (5)
}
\end{verbatim}
\end{tauenv}
It is noteworthy that we are relying on GPT to find evidences of synthetic and modified content, which is not ideal. Synthetic media detection is a specialized field locked in a constant "arms race" with generation techniques; dedicated detection models are often trained to spot specific artifacts that a general LLM may miss \cite{Akhtar2023DeepfakesShortSurvey, Thai2024DeepfakeDetectionDNN}. However, as we conduct the data analysis on the data, we notice that all testing points are real-world footage, thus have decided not to cumbersome the pipeline with another synthetic-detection model. In fact, a manual double-check survey on the final output of the pipeline shows no indications of AI-generated contents in the data.

\subsection{Putting together}
With all analysis information ready in Json format, we can easily and automatically assemble the report in any format. Our report strictly follows the requirement of the challenge. Furthermore, our metadata includes hyperlinks, allowing us to make references in our report.

\section{Pipeline Demonstration}
In this section, we demonstrate the output for test sample ``ID115'', and show that the pipeline produce a concurrent and cohesive response with explainable and controllable steps.

\noindent\textbf{Input.}
Two videos are provided with a description as below. We choose this sample for demonstration as it is violence-free.
\begin{tauenv}[frametitle=Sample ID115: Description]
\begin{verbatim}
{
    "location": "Supposedly Liman / Krasny Liman",
    "violence level": "(None) Military presence",
    "title": "",
    "media link": "https://t.me/zvezdanews/82025",
    "description": "During the liberation of Krasny 
    Liman, Russian soldiers found the nationalists" 
    commando post in the pioneer camp",
    "category": "Other"
 }
\end{verbatim}
\end{tauenv}
Henceforth, we are showing the output for each step.

\noindent\textbf{Step 1: From description to metadata}
\begin{tauenv}[frametitle=Search engine output for metadata]
\begin{verbatim}
[{"link": "https://english.ahram.org.eg/...",
  "date": "Not available",
  "title": "Russian army confirms capture of...",
  "content": "Failed to fetch the page."},
 {"link": "https://en.wikipedia.org/wiki/...",
  "date": "Not available",
  "title": "Battle of Krasnyi Lyman",
  "content": "..."},
 {"link": "https://tass.com/russia/1457247",
  "date": "May 28, 2022",
  "title": "Russia"s Armed Forces, DPR people...",
  "content": "..."},
 {"link": "https://en.iz.ru/en/1822471/...",
  "date": "Jan 15, 2025",
  "title": "Russian military began advancing...",
  "content": "..."},
 {"link": "https://tass.com/defense/1729357",
  "date": "Not available",
  "title": "Russian forces thwart four...",
  "content": "..."}, ...]
\end{verbatim}
\end{tauenv}

\noindent\textbf{Step 2: Videos \& Images Processing}
\begin{figure}[H]
    \centering
    \includegraphics[width=\linewidth]{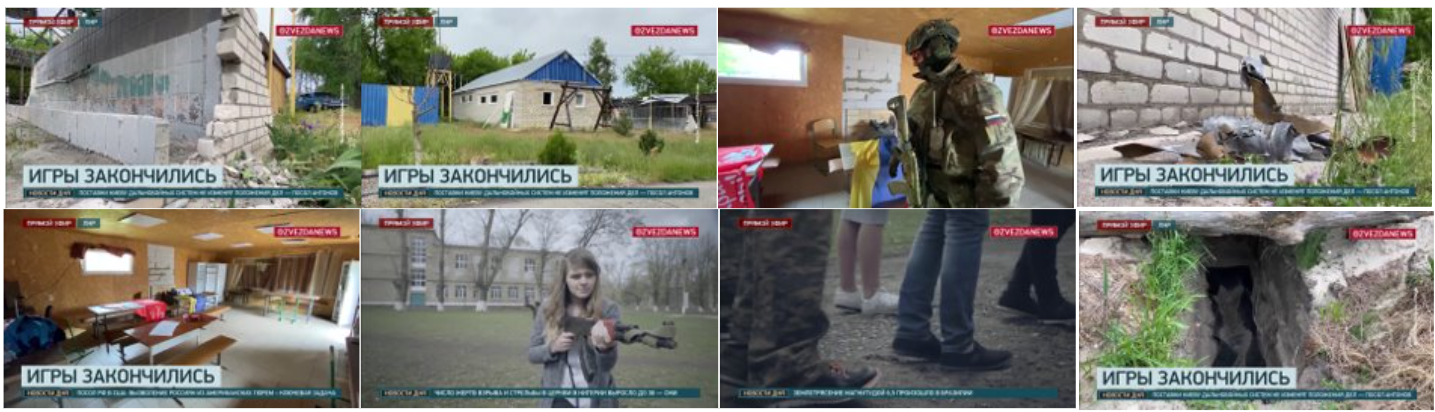}
    \caption{Selected Image Frames.}
    \label{fig:demo-image}
\end{figure}

\noindent\textbf{Step 3: Audio Processing}\\
\noindent Audio transcript is in Russian, which may or may not be fully cohesive depending on the quality of the audio itself. Here we omit showing the sample transcription due to technical difficulty in rendering Russian characters on \LaTeX.

\noindent\textbf{Step 4: Cross Validation of Information Sources}
\begin{tauenv}[frametitle=Cross Validation Output]
\begin{verbatim}
{
 "location": {
    "location": "Lyman (formerly Krasnyi Lyman), 
                Donetsk Oblast, Ukraine",
    "coordinates": "48.9781° N, 37.8017° E"
    },
 "date": {
    "date": "28/05/2022 - 02/10/2022",
    "concensus": "Partial",
    "notes": "First reported Russian capture: 
    28/05/2022 (TASS, TheJournal.ie). 
    Ukrainian recapture: 02/10/2022 (BBC). 
    Other sources reference ongoing fighting 
    in the area into 2023 and 2025, but the main
    events are concentrated between late May 
    and early October 2022."
        },
 "about": {
    "consensus": "The event concerns the battle 
    for control of the town of Lyman (formerly 
    Krasnyi Lyman) in Donetsk Oblast, Ukraine, 
    during the ongoing Russia-Ukraine war. 
    Russian and Donetsk People"s Republic (DPR) 
    forces captured the town from Ukrainian forces 
    in late May 2022, using it as a logistics hub. 
    Ukrainian forces recaptured the town in early 
    October 2022, marking a significant setback 
    for Russian operations in the region.",
 "conflicts": "Some sources (e.g., TASS, Russian 
    media) frame the Russian capture as a 
    "liberation" from "Ukrainian nationalists," 
    while Western and Ukrainian sources 
    (e.g., BBC) describe the Russian withdrawal 
    as a defeat and the Ukrainian recapture 
    as a liberation. There are also references 
    to ongoing fighting in the area after 
    October 2022, suggesting continued military 
    activity and contested control."},
 "tag": [
    "Ukraine War","Lyman","Donetsk","Russian 
    Armed Forces","Ukrainian Armed Forces","DPR"]
 }
\end{verbatim}
\end{tauenv}

\noindent\textbf{Step 5: Forensic Analysis}
\begin{tauenv}[frametitle=Forensic Analysis Output]
\begin{verbatim}
{
 "metadata-validation": {
    "location": "The images show a semi-rural or 
    peri-urban area with overgrown grass, a simple 
    brick building with a blue roof, and a fence 
    painted in the colors of the Ukrainian flag 
    (blue and yellow). The presence of Ukrainian 
    symbols and Cyrillic writing is consistent with 
    eastern Ukraine, specifically Donetsk Oblast. 
    The metadata states "Lyman (formerly Krasnyi 
    Lyman), Donetsk Oblast, Ukraine" as the location. 
    The visual evidence (Ukrainian flag colors, 
    Cyrillic documents, and the general environment) 
    supports this claim, though there is no direct 
    signage confirming "Lyman" in the images.",
    "event": "The metadata describes the event as 
    the capture of Lyman by Russian and DPR forces 
    in May 2022. The images show Russian military 
    presence (soldier with Russian flag patch), 
    abandoned Ukrainian symbols (flag, documents in 
    Ukrainian), and a disrupted environment, which 
    is consistent with a recent change of control. 
    The presence of Russian media (@zvezdanews) and 
    the phrase "The games are over" further support 
    the narrative of a takeover or end of Ukrainian 
    control.",
    "people": "Key individuals/groups identified: 
    Russian military personnel (visible in uniform 
    with Russian flag patch), possible Ukrainian 
    presence inferred from abandoned flags and 
    documents, and civilians (legs/feet visible in 
    one image, possibly detainees or locals)."
    },
 "authenticity": "The content appears authentic and 
    not synthetic or AI-generated. The images are 
    consistent with field reporting and do not show 
    signs of digital fabrication.",
 "auth-evidence": "Evidence for authenticity 
    includes: natural lighting and shadows, 
    realistic details in the environment, 
    consistent camera angles and quality, and 
    the presence of real-world objects (documents, 
    flags, uniforms) that match the context. 
    The documents are in Ukrainian, and the 
    flags and uniforms are consistent with the 
    reported parties.",
 "synt-type": "No evidence of AI-generated content 
    (GANs, Stable Diffusion, etc.) or synthetic 
    imagery. The images appear to be original 
    field footage.",
 "other": "No major anomalies or manipulations 
    detected. The only notable aspect is the 
    media source (@zvezdanews), which is a 
    Russian state outlet, so editorial framing 
    should be considered, but the visual content 
    itself does not show signs of digital 
    alteration."
}
\end{verbatim}
\end{tauenv}

% \section{Report on Validation dataset}

\section{Conclusion}
In this work, we present a pipeline for information verification as a solution for the ACMMM'25 grand challenge. The pipeline consists of three main components: metadata retrieval, audio processing, and cross validation \& forensic analysis. Powered by AI and prompt engineering, we design a fully automated pipeline with minimal human interventions. 

\subsection{Limitations}
There are several drawbacks regarding the workflow of our proposal pipeline. Firstly, our pipeline functions solely on text-based metadata, which limits our capacity to retrieve internet information from multi-media platforms such as Twitter (now X) or Youtube. Secondly, for resources reasons, we are currently rely on GPT knowledge base for geographical information such as roads, buildings, and cities. However, as far as GPT can provide, it is unconvincing that accurate coordinates can be extracted without using specific tools such as Google Map. Finally, we omit a dedicated AI-content detection component, as the testing data provided does not show evidence of such content. This was a strategic choice for the challenge, but broader news verification requires it \cite{Pei2024DeepfakeBenchmarkSurvey}. A more robust system could integrate proactive provenance technologies like C2PA's Content Credentials \cite{C2PA2025ContentCredentials} or watermarking like SynthID \cite{Kohli2025SynthID}, which certify authentic content at its source rather than just reactively detecting fakes.

\subsection{Discussion}
The task to verify information is challenging, even for humans, requiring extensive data gathering and cross-validation. While our pipeline addresses these, its results might not reach human level due to the aforementioned drawbacks. Future work should focus on: (i) a multi-modal metadata retrieval system; (ii) incorporating specific tools like Google Map; (iii) updating the backbone LLM \cite{Pan2024AssessingImproving}; (iv) improving model generalization across news domains \cite{Hamed2023FakeNewsReview}; and (v) tackling the "dataset dilemma" by developing high-quality, diverse, datasets \cite{Alsudais2024FakeNewsDatasetSurvey}.

\printbibliography

%----------------------------------------------------------

\end{document}